\documentclass[11pt,a4paper]{article}
\usepackage[hyperref]{acl2020}
\usepackage{times,latexsym}
\usepackage{url}
\usepackage[T1]{fontenc}
\usepackage{graphicx}
\usepackage{stmaryrd}
\usepackage{amsmath}
\usepackage{amsfonts}
\usepackage{colortbl}

\usepackage{booktabs}
\usepackage{lingmacros}

\newcommand{\fm}[1]{\textsl{#1}}
\newcommand{\cut}[1]{}

\newcommand{\exam}[1]{\textit{#1}}

\usepackage{microtype}

\aclfinalcopy 
\def\aclpaperid{2539} 

\title{Probing Linguistic Systematicity}

\author{Emily Goodwin,\thanks{~~Corresponding author} \textsuperscript{1,5}
  Koustuv Sinha,\textsuperscript{2,3,5}
  Timothy J. O'Donnell\textsuperscript{1,4,5} \\ 
  \textsuperscript{1}Department of Linguistics,
  \textsuperscript{2}School of Computer Science, ~~McGill University,
  Canada \\
  \textsuperscript{3}Facebook AI Research (FAIR), Montreal
  ~~~~~~\textsuperscript{4}Canada CIFAR AI Chair, Mila\\
     \textsuperscript{5}Quebec Artificial Intelligence Institute (Mila) \\
     \texttt{\normalsize \{emily.goodwin, koustuv.sinha\}@mail.mcgill.ca} \\
     \texttt{\normalsize timothy.odonnell@mcgill.ca}
  \\
}

\date{}

\begin{document}
\maketitle
\begin{abstract}
  Recently, there has been much interest in the question of whether deep natural language understanding models exhibit
  \textit{systematicity}---generalizing such that units like words make consistent contributions to the meaning of the sentences in which they appear.
    There is accumulating evidence that
  neural models often generalize non-systematically. We examined the notion of
  systematicity from a linguistic perspective, defining a set of probes
  and a set of metrics to measure systematic behaviour. We also
  identified ways in which network architectures can generalize
  non-systematically, and discuss why such forms of generalization may
  be unsatisfying. As a case study, we performed a series of experiments
  in the setting of natural language inference (NLI), demonstrating
  that some NLU systems achieve high overall performance
  despite being non-systematic. 

\end{abstract}

\section{Introduction}
\label{sect:intro}


 Language allows us to express and
comprehend a vast variety of novel thoughts and ideas. This creativity
is made possible by \fm{compositionality}---the linguistic system
builds utterances by combining an inventory of primitive units such as
morphemes, words, or idioms (the \fm{lexicon}), using a small set of
structure-building operations \citep[the
\fm{grammar};][]{carnap1947meaning,fodor1988,hodges2012formalizing,
  janssen2012compositionality,lake2017building,szabo2012case,
  zadrozny1994compositional, lake2019human}.

One property of compositional systems, widely studied in the cognitive sciences,
is the phenomenon of
\fm{systematicity}. Systematicity refers to the fact that lexical
units such as words make consistent contributions to the meaning of
the sentences in which they appear. \citet{fodor1988} provided a famous
example: If a competent speaker of English knows the meaning of the
sentence \exam{John loves the girl}, they also know 
the meaning of \exam{The girl loves John}. This is because for
speakers of English knowing the meaning of the first sentence implies
knowing the meaning of the individual words \exam{the}, \exam{loves},
\exam{girl}, and \exam{John} as well as grammatical principles such as how
transitive verbs take their arguments. But knowing these words and
principles of grammar implies knowing how to compose the meaning of
the second sentence.

Deep learning systems now regularly exhibit very high performance
on a large variety of natural language tasks, including machine
translation \citep{wu2016googles, vaswani2017attention}, question
answering \citep{wang2018multi, henaff2016tracking}, visual question
answering \citep{hudson2018compositional}, and natural language
inference \citep{devlin2018bert, storks2019recent}.  Recently,
however, researchers have asked whether such systems generalize
systematically (see \textsection\ref{sec:related}). 
  
Systematicity is the property whereby words have consistent
contributions to composed meaning; the alternative is the situation
where words have a high degree of \fm{contextually conditioned meaning
  variation}. In such cases, generalization may be based on \fm{local
  heuristics} \citep[][]{mccoy.r:2019, niven2019probing},
\fm{variegated similarity} \citep[][]{albright.a:2003}, or \fm{local
  approximations} \citep{veldhoen2017recursive}, where the
contribution of individual units to the meaning of the sentence can
vary greatly across sentences, interacting with other units in highly
inconsistent and complex ways.

This paper introduces several novel probes for testing systematic generalization.
We employ an artificial language to have control over systematicity and contextual meaning variation. 
Applying our probes to this language in an NLI setting reveals that some deep learning systems which achieve very high accuracy on standard holdout evaluations do so in ways which are non-systematic: the networks do not consistently capture the basic notion that certain classes of words have meanings which are
consistent across the contexts in which they appear.

The rest of the paper is organized as follows. 
\textsection\ref{sect:SystematicityandCompositionality} discusses
degrees of systematicity and contextually conditioned variation;
\textsection\ref{sect:CompositionalStructureinNaturalLanguage}
introduces the distinction between open- and closed-class words, which
we use in our probes.  \textsection\ref{sect:studysetup} introduces
the NLI task and describes the artificial language we use;
\textsection\ref{sect:simulations} discusses the models that we tested
and the details of our training setup; \textsection\ref{sect:probingSystematicity}
introduces our probes of systematicity and results are presented in
\textsection\ref{sect:analyses}.\footnote{Code for datasets and models
  can be found here: \href{https://github.com/emilygoodwin/systematicity}{https://github.com/emilygoodwin/systematicity}}

\section{Systematicity and Contextual Conditioning}
\label{sect:SystematicityandCompositionality}

Compositionality is often stated as the principle that the meaning of
an utterance is determined by the meanings of its parts and
the way those parts are combined \citep[see,
e.g.,][]{heim.i:2000}.

Systematicity, the property that words mean the same thing in
different contexts, is closely related to compositionality;
nevertheless, compositional systems can vary in their degree of
systematicity. At one end of the spectrum are systems in which
primitive units contribute exactly one identical meaning across all
contexts. This high degree of systematicity is approached by
artificial formal systems including programming languages and logics,
though even these systems don't fully achieve this ideal
\citep[][]{cantwell-smith.b:1996,dutilh-novaes.c:2012}.

The opposite of systematicity is the phenomenon of \fm{contextually
  conditioned variation in meaning} where the contribution of
individual words varies according to the sentential contexts in which
they appear. Natural languages exhibit such context dependence in
phenomena like homophony, polysemy, multi-word idioms, and
co-compositionality. Nevertheless, there are many words in natural
language---especially closed-class words like quantifiers (see
below)---which exhibit very little variability in meaning across
sentences.

At the other end of the spectrum from programming languages and logics
are systems where many or most meanings are highly context
dependent. The logical extreme---a system where each word has a
different and unrelated meaning every time it occurs---is clearly of
limited usefulness since it would make generalization
impossible. Nevertheless, learners with sufficient memory capacity and
flexibility of representation, such as deep learning models, can learn
systems with very high degrees of contextual conditioning---in
particular, higher than human language learners.  An important goal
for building systems that learn and generalize like people is to
engineer systems with inductive biases for the right degree of
systematicity.  In \textsection\ref{sect:analyses}, we give evidence
that some neural systems are likely too biased toward allowing
contextually conditioned meaning variability for words, such as
quantifiers, which do not vary greatly in natural language.

\section{Compositional Structure in Natural Language}
\label{sect:CompositionalStructureinNaturalLanguage}

Natural language distinguishes between
\fm{content} or \fm{open-class} lexical units and \fm{function} or
\fm{closed-class} lexical units. The former refers to
categories, such a nouns and verbs, which carry the majority of
contentful meaning in a sentence and which permit new
coinages. Closed-class units, by contrast, carry most of the grammatical structure of the sentence and consist of things like
inflectional morphemes (like pluralizing \exam{-s} in English) and
words like determiners, quantifiers, and negation (e.g., \exam{all, some, the} in English). These are mostly fixed; adult speakers do not coin new quantifiers, for example, the way that they coin new nouns. 

Leveraging this distinction gives rise to the possibility of
constructing probes based on \fm{jabberwocky-type sentences}. This
term references the poem \textit{Jabberwocky} by Lewis Carroll, which
combines nonsense open-class words with familiar closed-class words in
a way that allows speakers to recognize the expression as well formed. For example, English speakers identify a contradiction in the sentence \exam{All Jabberwocks flug, but some Jabberwocks don't flug}, 
without a meaning for \exam{jabberwock} and \exam{flug}. This is
possible because we expect the words \exam{all},
\exam{some}, \exam{but}, and \exam{don't} to contribute the same
meaning as they do when combined with familiar words, like \exam{All
  pigs sleep, but some pigs don't sleep}.

Using jabberwocky-type sentences, we tested the generalizability of
certain closed-class word representations learned by neural networks.
Giving the networks many examples of each construction with a large
variety of different content words---that is, large amounts of highly
varied evidence about the meaning of the closed-class words---we asked
during the test phase how fragile this knowledge is when transferred
to new open-class words. That is,
our probes combine novel open-class words with familiar closed-class
words, to test whether the closed-class words are treated
systematically by the network. For example, we might train
the networks to identify contradictions in pairs like \exam{All pigs
  sleep; some pigs don't sleep}, and test whether the network can
identify the contradiction in a pair like \exam{All Jabberwocks flug;
  some Jabberwocks don't flug}. A systematic learner would reliably
identify the contradiction, whereas a non-systematic learner may allow
the closed-class words (\exam{all, some, don't}) to take on contextually conditioned meanings
that depend on the novel context words.

\section{Related Work}
\label{sec:related}
There has been much interest in the problem of systematic
generalization in recent years \citep[][inter alia]{
bahdanau.d:2019,bentivogli2016sick,lake2019human,lake2017building,gershman2015,mccoy2019berts,veldhoen2017recursive,soulos2019discovering,prasad2019using,richardson2019probing, johnson2017clevr}.

In contrast to our approach (testing novel words in familiar
combinations), many of these studies probe systematicity by testing
familiar words in novel combinations.  \citet{lake2018generalization}
adopt this approach in semantic parsing with an artificial language
known as SCAN. \citet{dasgupta2018evaluating, dasgupta2019analyzing}
introduce a naturalistic NLI dataset, with test items that shuffle the
argument structure of natural language utterances. In the in the
inductive logic programming domain, \citet{sinha2019clutrr} introduced
the CLUTTR relational-reasoning benchmark. The
novel-combinations-of-familiar-words approach was formalized in the
CFQ dataset and associated distribution metric of
\citet{keysers2019measuring}. \citet{ettinger2018assessing} introduced
a semantic-role-labeling and negation-scope labeling dataset, which
tests compositional generalization with novel combinations of familiar
words and makes use of syntactic constructions like relative clauses.
Finally, \citet{kim2019probing} explore pre-training schemes'
abilities to learn prepositions and wh-words with syntactic
transformations (two kinds of closed-class words which our work does
not address).

A different type of systematicity analysis directly investigates
learned representations, rather than developing probes of model
behavior. This is done either through visualization
\citep{veldhoen2017recursive}, training a second network to
approximate learned representations using a symbolic structure
\citep{soulos2019discovering} or as a \fm{diagnostic classifier}
\citep[][]{giulianelli2018under}, or reconstructing the semantic space
through similarity measurements over representations
\citep{prasad2019using}.

\section{Study Setup}
\label{sect:studysetup}
\subsection{Natural Language Inference}
We make use of the \fm{Natural language inference} (NLI) task to study the question of
systematicity. 
The NLI task is to infer the relation between two
sentences (the \fm{premise} and the \fm{hypothesis}). Sentence pairs must be classified into one of a set of predefined
logical relations such as \fm{entailment} or \fm{contradiction}. For
example, the sentence \exam{All mammals growl} entails the sentence
\exam{All pigs growl}.  A rapidly growing number of studies have shown
that deep learning models can achieve very high performance in this
setting \citep{evans2018can, conneau2017supervised,
  bowman2014recursive, yoon2018dynamic, kiela2018dynamic,
  munkhdalai2017neural, rocktaschel2015reasoning, peters2018deep,
  parikh2016decomposable, zhang2018know,
  radford2018improving, devlin2018bert, storks2019recent}.
\subsection{Natural Logic}
\label{sect:natlog}
We adopt the formulation of NLI known as \fm{natural logic}
\citep[][]{MacCartney2014,MacCartney:2009:EMN,lakoff1970linguistics}. Natural
logic makes use of seven logical \fm{relations} between pairs of
sentences. These are shown in Table~\ref{tbl:natlog}. These relations
can be interpreted as the set theoretic
relationship between the extensions of the two expressions. For
instance, if the expressions are the simple nouns \exam{warthog} and
\exam{pig}, then the entailment relation ($\sqsubset$) holds between
these extensions (\exam{warthog} $\sqsubset$ \exam{pig}) since every
warthog is a kind of pig.

For higher-order operators such as quantifiers, relations can be
defined between sets of possible worlds. For instance, the set of
possible worlds consistent with the expression \exam{All blickets wug}
is a subset of the set of possible worlds consistent with the
logically weaker expression \exam{All red blickets wug}. Critically,
the relationship between composed expressions such as \exam{All X Y}
and \exam{All P Q} is determined entirely by the relations between X/Y
and P/Q, respectively. Thus, natural logic allows us to compute the
relation between the whole expressions using the relations between
parts. We define an artificial language in which such alignments are
easy to compute, and use this language to probe deep learning systems'
ability to generalize systematically.
\begin{table}[ht]
\centering
\resizebox{\linewidth}{!}{
\begin{tabular}{lllll}
\toprule
Symbol & Name & Example & Set-theoretic definition \\
\midrule
\rowcolor{gray!6} $x \equiv y$ & equivalence & $pig \equiv pig$ & $x = y$ \\
$x \sqsubset y$ & forward entailment & $pig \sqsubset mammal$ & $x \subset y$  \\
\rowcolor{gray!6} $x \sqsupset y$ & reverse entailment & $mammal \sqsupset pig$ & $x \supset y $\\
$x\ ^\wedge\ y$ & negation & $pig\ ^\wedge\ not \,  pig$ & $x \cap y  = \emptyset \wedge x \cup y = U$ \\
\rowcolor{gray!6} $x \mid y$ & alternation & $pig \mid cat$ & $x \cap y  = \emptyset \wedge x \cup y \neq U $ \\
$x \smile y$ & cover & $ mammal \smile not \,  pig $ & $x \cap y  \neq \emptyset \wedge x \cup y = U
$ \\
\rowcolor{gray!6} $x \# y$ & independence & $hungry\,\#\,warthog$ & (all other cases) \\
\bottomrule
\end{tabular}}
\caption{\citet{MacCartney:2009:EMN}'s implementation of natural logic relations}\label{tbl:natlog}
\end{table}

\subsection{The Artificial Language}
\label{sect:thelanguage}

In our artificial language, sentences are generated according to the
six-position template shown in Table~\ref{tbl:artlang}, and include a quantifier (position 1), noun (position 3), and verb
(position 6), with optional pre- and post-modifiers (position 2 and
4) and optional negation (position 5). For readability, all examples
in this paper use real English words; however, simulations can use uniquely identified abstract symbols (i.e., generated by \texttt{gensym}).

\begin{table*}[t]
  \centering
  \resizebox{\textwidth}{!}{
    \begin{tabular}{|l||llllll|}
      \hline
        \textbf{Position} & 1 & 2 & 3 &  4 & 5 & 6 \\ \hline
  \rowcolor{gray!6} \textbf{Category} & \texttt{quantifier} & \texttt{nominal premodifier} & \texttt{noun} &  \texttt{nominal postmodifier} & \texttt{negation} & \texttt{verb}  \\
  \textbf{Status} & Obligatory & Optional & Obligatory &  Optional & Optional & Obligatory\\
  \rowcolor{gray!6} \textbf{Class} & Closed & Closed & Open & Closed & Closed & Open \\
  \textbf{Example} & {\Large \exam{All}} &  {\Large \exam{brown}} &  {\Large  \exam{dogs}} &  {\Large  \exam{that bark}} &  {\Large \exam{don't}} &  {\Large \exam{run}} \\ 
  \hline
\end{tabular}
}
\caption{A template for sentences in the artificial language. Each sentence fills the obligatory positions 1, 3, and 6 with a word: a quantifier, noun, and verb. Optional positions (2, 4 and 5) are filled by either a word (adjective, postmodifier or negation) or by the empty string. Closed-class categories (Quantifiers, adjectives, post modifiers and negation) do not include novel words, while  open-class categories (nouns and verbs) includes novel words that are only exposed in the test set.
} \label{tbl:artlang}
\end{table*}

We compute the relation between position-aligned pairs of sentences in
our language using the natural logic system (described in
\textsection\ref{sect:natlog}). Quantifiers and negation have their usual 
natural-language semantics in our artificial language; pre- and post-modifiers are
treated intersectively. Open-class items (nouns and verbs) are
organized into linear hierarchical taxonomies, where each
open-class word is the sub- or super-set of exactly one other
open-class item in the same taxonomy. For example, since \exam{dogs}
are all \exam{mammals}, and all mammals \exam{animals}, they form the
entailment hierarchy \exam{dogs} $\sqsubset$ \exam{mammals}
$\sqsubset$ \exam{animals}. We vary the number of distinct noun and verb taxonomies
according to an approach we refer to as \fm{block} structure, described in
the next section.

\subsection{Block Structure}
\label{sect:blockstructure}

In natural language, most open-class words do not appear with equal
probability with every other word. Instead, their distribution
is biased and clumpy, with words in similar
topics occurring together. To mimic such topic structure, we group
nouns and verbs into \fm{blocks}.
Each block consists of six nouns
and six verbs, which form taxonomic hierarchies (e.g.,
\exam{lizards}/\exam{animals}, \exam{run}/\exam{move}).  Nouns and
verbs from different blocks have no taxonomic relationship (e.g.,
\exam{lizards} and \exam{screwdrivers} or \exam{run} and \exam{read})
and do not co-occur in the same sentence pair. Because each
block includes a six verbs and six nouns in a linear taxonomic hierarchy, no single block is
intrinsically harder to learn than any other block.

The same set of closed-class words appear with all blocks of open-class words, 
and their meanings are systematic regardless of the open-class words (nouns and verbs)
they are combined with. For example, the quantifier \exam{some} has a consistent 
meaning when it is applied to \exam{some screwdrivers} or \exam{some animals}.
Because closed-class words are shared across blocks, models are trained on extensive and varied evidence of their behaviour. We present closed-class words in a wide variety 
of sentential contexts, with a wide variety of different open-class words, to
provide maximal pressure against overfitting and maximal evidence of their consistent meaning.

\subsection{Test and Train Structure}
\label{sect:training-setup}
We now describe the structure of our training blocks,
holdout test set, and jabberwocky blocks. We also discuss our two test
conditions, and several other issues that arise in the construction of
our dataset.

\paragraph{Training set:} For
each training block, we sampled (without replacement) one sentence pair for
every possible combination of open-class words, that is, every
combination of nouns and verbs
$\langle \mathrm{\texttt{noun}}_1, \mathrm{\texttt{noun}}_2,
\mathrm{\texttt{verb}}_1,
\mathrm{\texttt{verb}}_2\rangle$. Closed-class words were sampled
uniformly to fill each remaining positions in the sentence (see
Table~\ref{tbl:artlang}). A random subset of
20\% of training items were reserved for validation (early stopping)
and not used during training.

\paragraph{Holdout test set:} For each training block, we sampled a
holdout set of forms using the same nouns and verbs, but disjoint from
the training set just described. The sampling procedure was identical
to that for the training blocks. These holdout items allow us to test
the generalization of the models with known words in novel
configurations (see \textsection\ref{sec:holdout}). 

\paragraph{Jabberwocky test set:} Each \fm{jabberwocky block} consisted of
novel open-class items (i.e., nouns and verbs) that did not appear in
training blocks. For each jabberwocky block, we began by following a
sampling procedure identical to that for the training/holdout sets
with these new words. Several of our systematicity probes are
based on the behavior of neighboring pairs of test sentences (see
\textsection\ref{sect:probingSystematicity}). To ensure that all
such necessary pairs were in the jabberwocky test set, we extended the
initial sample with any missing test items.

\paragraph{Training conditions:}
Since a single set of closed-class words is used across all blocks,
adding more blocks increases evidence of the meaning of these words
without encouraging overfitting. To study the effect of increasing
evidence in this manner, we use two training conditions: \fm{small}
with 20 training blocks and \fm{large} with 185 training blocks. Both conditions contained 20 jabberwocky blocks.
The small condition consisted of $51,743$ training, $10,399$
validation, and $3,694,005$ test (holdout and jabberwocky) pairs. The
large condition consisted of $478,649$ training, $96,005$ validation,
and $3,694,455$ test items.

\paragraph{Balancing:}
One consequence of the sampling method is that logical relations will
not be equally represented in training. In fact, it is impossible to
simultaneously balance the distributions of syntactic constructions,
logical relations, and instances of words. In this trade-off, we chose
to balance the distribution of open-class words in the vocabulary, as
we are focused primarily on the ability of neural networks to
generalize closed-class word meaning. Balancing instances of open-class words  provided the greatest variety of learning contexts for the meanings of the closed-class items.

\section{Simulations}
\label{sect:simulations}
\subsection{Models}
\label{sec:models}

We analyze performance on four simple baseline models known to perform
well on standard NLI tasks, such as the Stanford Natural Language
Inference datasets, \citep{bowman2015large}. Following
\citet{conneau2017supervised}, the hypothesis $u$ and premise $v$ are
individually encoded by neural sequence encoders such as a long
short-term memory \citep[LSTM;][]{hochreiter1997long} or gated
recurrent unit \citep[GRU;][]{cho2014properties}. These vectors,
together with their element-wise product $u*v$ and element-wise
difference $u-v$ are fed into a fully connected multilayer perceptron
layer to predict the relation. The encodings $u$ and $v$ are produced
from an input sentence of $M$ words, $w_1, \ldots, w_M$, using a
recurrent neural network, which produces a set of a set of $M$ hidden
representations $h_1, \ldots, h_t$, where $h_t = f(w_1, \ldots, w_M)$.
The sequence encoding is represented by its last hidden vector $h_T$.

The simplest of four models sets $f$ to be a bidirectional gated
recurrent unit (BGRU). This model concatenates the last hidden state
of a GRU run forwards over the sequence and the last hidden state of
GRU run backwards over the sequence, for example,
$u = [\overleftarrow{h_M}, \overrightarrow{h_M}]$.

Our second embedding system is the Infersent model reported by
\citet{conneau2017supervised}, a bidirectional LSTM with max pooling
(INFS). This is a model
where $f$ is an LSTM. Each word is represented by the concatenation of
a forward and backward representation:
$h_t = [\overleftarrow{h_t}, \overrightarrow{h_t}]$. We constructed a
fixed vector representation of the sequence $h_t$ by selecting the
maximum value over each dimension of the hidden units of the words in
the sentence.

Our third model is a self-attentive sentence encoder (SATT) which uses
an attention mechanism over the hidden states of a BiLSTM to generate
the sentence representation \cite{lin2017structured}. This attention
mechanism is a weighted linear combination of the word
representations, denoted by $u = \sum_{M} \alpha_i h_i$, where the
weights are calculated as follows:
\begin{equation*}
    \bar{h_i} = \text{tanh}(Wh_i + b_w)
\end{equation*}
\begin{equation*}
    \alpha_i = \frac{e^{\bar{h_i}^\top u_w}}{\sum_i e^{\bar{h_i}^\top u_w}}
\end{equation*}
\noindent where, $u_w$ is a learned context query vector and
$(W, b_w)$ are the weights of an affine transformation. This self-attentive network also has multiple views of the sentence, so the
model can attend to multiple parts of the given sentence at the same
time.

Finally, we test the Hierarchical ConvolutionalNetwork (CONV)
architecture from \cite{conneau2017supervised} which is itself
inspired from the model \textit{AdaSent} \cite{zhao2015self}. This
model has four convolution layers; at each layer the intermediate
representation $u_i$ is computed by a max-pooling operation over
feature maps. The final representation is a concatenation
$u = [u_1, ... , u_l]$ where $l$ is the number of layers. 

\section{Probing Systematicity}
\label{sect:probingSystematicity}

In this section, we study the systematicity of the models described in
\textsection\ref{sec:models}. Recall that systematicity refers to the
degree to which words have consistent meaning across different
contexts, and is contrasted with contextually conditioned
variation in meaning. We describe three novel probes of systematicity which we
call the \fm{known word perturbation probe}, the \fm{identical open-class words
  probe}, and the \fm{consistency probe}.

All probes take advantage of the distinction between closed-class and
open-class words reflected in the design of our artificial language, and are performed on sentence pairs with novel open-class words
  (jabberwocky-type sentences; see \textsection\ref{sect:training-setup}
). We now describe the logic of each
probe.

\subsection{Known Word Perturbation Probe}
\label{sec:known_word_probe}

We test whether the models treat the meaning of closed-class words
systematically by perturbing correctly classified jabberwocky sentence
pairs with a closed-class word.  More precisely, for a pair of 
closed-class words $w$ and $w^\prime$, we consider test items which
can be formed by substitution of $w$ by $w^\prime$ in a correctly
classified test item. We allow both $w$ and $w^\prime$ to be any of
the closed-class items, including quantifiers, negation, nominal post-modifiers, 
or the the empty string $\epsilon$ (thus modeling insertions and deletions of
these known, closed-class items). Suppose that Example~\ref{ex:transform1}
was correctly classified. Substituting \exam{some} for \exam{all} in
the premise of yields Example \ref{ex:transform2}, and changes the
relation from entailment ($\sqsubset$) to reverse entailment
($\sqsupset$).

\enumsentence{\label{ex:transform1}\textbf{All} blickets wug.\\All blockets wug.}

\enumsentence{\label{ex:transform2}\textbf{Some} blickets wug.\\All 
  blockets wug.}
  
There are two critical features of this probe. First, because we start
from a correctly-classified jabberwocky pair,
we can conclude that the novel words (e.g., \exam{wug} and
\exam{blickets} above) were assigned appropriate meanings. 

Second, since the perturbation only involves closed-class items which
do not vary in meaning and have been highly trained, the perturbation
should not affect the models ability to correctly classify the
resulting sentence pair. If the model does misclassify the resulting
pair, it can only be because a perturbed closed-class word (e.g.,
\exam{some}) interacts with the open-class items (e.g., \exam{wug}),
in a way that is different from the pre-perturbation closed-class item
(i.e., \exam{all}). This is non-systematic behavior.

In order to rule out trivially correct behavior where the model simply
ignores the perturbation, we consider only perturbations which result
in a change of class (e.g., $\sqsubset \mapsto \sqsupset$) for the
sentence pair.  In addition to accuracy on these perturbed items, we also
examine the variance of model accuracy on probes across different
blocks. If a model's accuracy varies depending only on the novel
open-class items in a particular block, this provides further evidence
that it does not treat word meaning systematically.

\subsection{Identical Open-class Words Probe}
\label{sec:lexical_id_probe}

Some sentence pairs are classifiable without any knowledge of the
novel words' meaning; for example, pairs where
premise and hypothesis have identical open-class words.  An instance
is shown in Example~\ref{ex:negation}: the two
sentences must stand in contradiction, regardless of the meaning of
\textit{blicket} or \textit{wug}.

\enumsentence{\label{ex:negation}All blickets wug.\\Some blickets don't
  wug.}

The closed-class items and compositional structure of the language is
sufficient for a learner to deduce the relationships between such
sentences, even with unfamiliar nouns and verbs. Our second probe, the
\fm{identical open-class words probe}, tests the models' ability to correctly
classify such pairs. 

\subsection{Consistency Probe}

Consider Examples~\ref{ex:entailment} and ~\ref{ex:reverse-entailment}, which present the same two sentences in opposite orders. 

\enumsentence{\label{ex:entailment}All blickets wug.\\All red blickets
  wug.}  \enumsentence{\label{ex:reverse-entailment}All red blickets
  wug.\\All blickets wug.}

In Example~\ref{ex:entailment}, the two sentences stand in an
entailment ($\sqsubset$) relation. In
Example~\ref{ex:reverse-entailment}, by contrast, the two sentences
stand in a reverse entailment ($\sqsupset$) relation. This is a
logically necessary consequence of the way the relations are defined. 
Reversing the order of sentences has predictable effects for all seven
natural logic relations: in particular, such reversals map
$\sqsubset \mapsto \sqsupset$ and $\sqsupset \mapsto \sqsubset$, leaving
all other relations intact. Based on this observation, we
develop a consistency probe of systematicity. We ask for each
correctly classified jabberwocky block test item, whether the corresponding
reversed item is also correctly classified. The intuition behind this
probe is that whatever meaning a model assumes for the novel
open-class words, it should assume the same meaning when the sentence
order is reversed. If the reverse is not correctly classified, then
this is strong evidence of contextual dependence in meaning.

\section{Results}
\label{sect:analyses}

In this section, we report the results of two control analyses, and that of our three systematicity probes described above. 
 
\subsection{Analysis I: Holdout Evaluations}
\label{sec:holdout}

We first establish that the models perform well on novel
configurations of known words. Table~\ref{tbl:holdout} reports
accuracy on heldout sentence pairs, described in
\textsection\ref{sect:training-setup}. The table reports average
accuracies across training blocks together with the standard
deviations of these statistics. As can be seen in the table, all
models perform quite well on holdout forms across training blocks,
with very little variance. Because these items use the
same sampling scheme and vocabulary as the trained blocks, these
simulations serve as a kind of upper bound on the performance and a
lower bound on the variance that we can expect from the more
challenging jabberwocky-block-based evaluations below.

\begin{table}[ht]
\centering
\resizebox{\linewidth}{!}{
\begin{tabular}{lllll}
\toprule
Condition & BGRU & CONV & SATT & INFS \\\
 & {\small mean (sd)} & {\small mean (sd)} & {\small mean (sd)} &{\small mean (sd)} \\
\midrule
small & 95.1 $\pm0.21$ & 95.43 $\pm0.12$ & 93.14 $\pm0.94$ & 96.02 $\pm0.51$ \\
large & 95.09 $\pm1.03$ & 95.22 $\pm0.55$ & 94.89 $\pm1.09$ & 96.17 $\pm0.74$ \\
\bottomrule
\end{tabular}}
\caption{Accuracy on holdout evaluations (training conditions and holdout evaluation are explained in  \textsection\ref{sect:training-setup})}\label{tbl:holdout}
\end{table}

\subsection{Analysis II: Distribution of Novel Words}
\label{sec:transformations}

\begin{figure}[!htb]
    \includegraphics[width=0.48\textwidth]{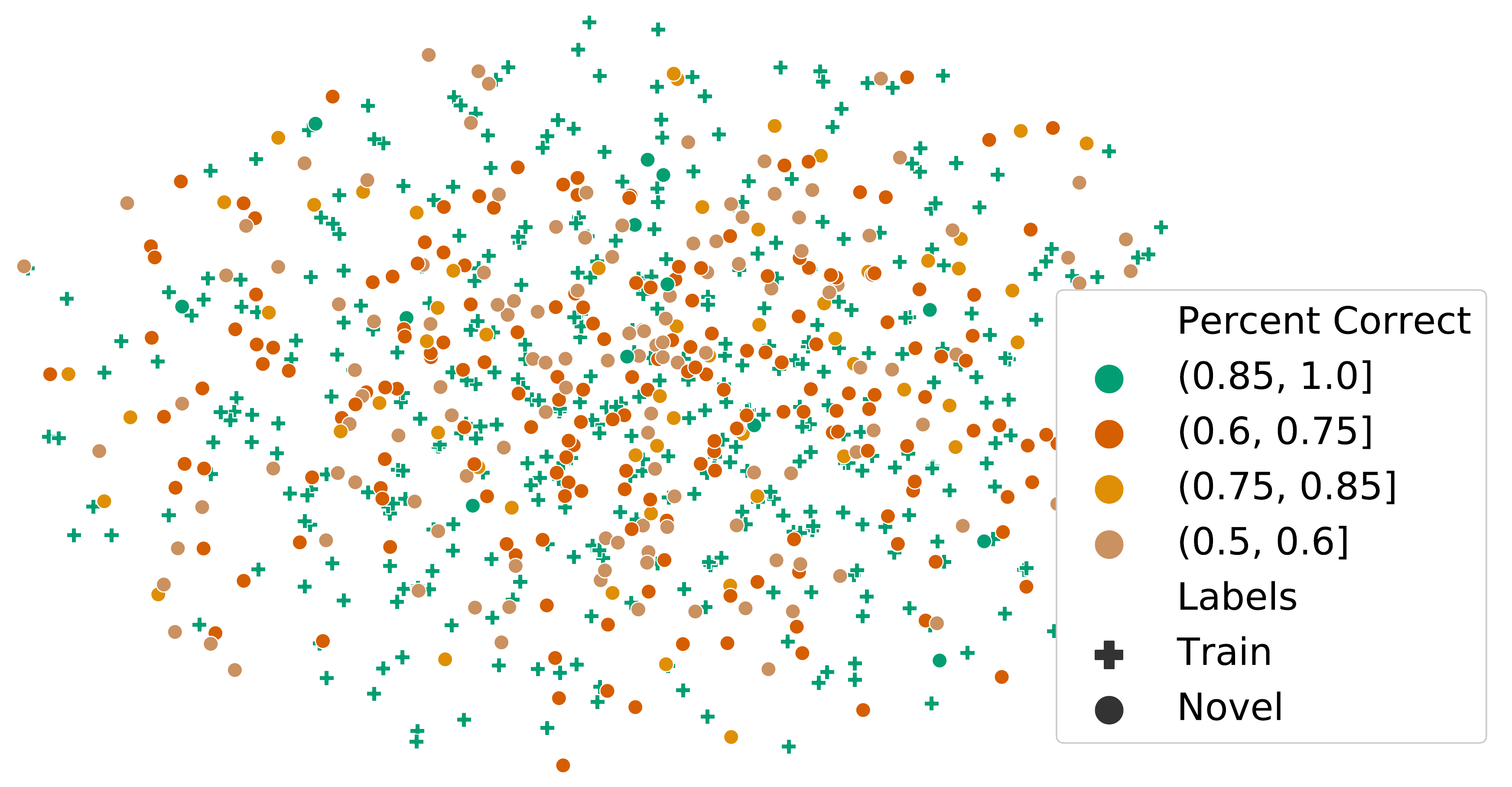}
    \caption{Visualization of trained and novel open-class word embeddings.}
    \label{fig:word_dist}
\end{figure}

Our three systematicity probes employ jabberwocky-type
sentences---novel open-class words in sentential frames built from
known closed-class words. Since models are not trained on these novel
words, it is important to establish that they are from the same
distribution as the trained words and, thus, that the models'
performance is not driven by some
pathological feature of the novel word embeddings.

Trained word embeddings were initialized randomly from
$\mathcal{N}(0,1)$ and then updated during training. Novel word
embeddings were simply drawn from $\mathcal{N}(0,1)$ and never
updated. Figure~\ref{fig:word_dist} plots visualizations of the
trained and novel open-class word embeddings in two dimensions, using
t-SNE parameters computed over all open-class words
\citep[][]{maaten2008visualizing}. Trained words are plotted as $+$,
novel words as $\bullet$. Color indicates the proportion of test items
containing that word that were classified correctly.  As the plot
shows, the two sets of embeddings overlap considerably.  Moreover,
there does not appear to be a systematic relationship between rates of
correct classification for items containing novel words and their
proximity to trained words.  We also performed a
resampling analysis, determining that novel vectors did not differ
significantly in length from trained vectors ($p=0.85$).  Finally, we
observed mean and standard deviation of the pairwise cosine similarity
between trained and novel words to be $0.999$ and $0.058$
respectively, confirming that there is little evidence the
distributions are different.

\subsection{Analysis III: Known Word Perturbation Probe}
\label{sec:anal3}

\begin{figure}
	\includegraphics[width = \linewidth]{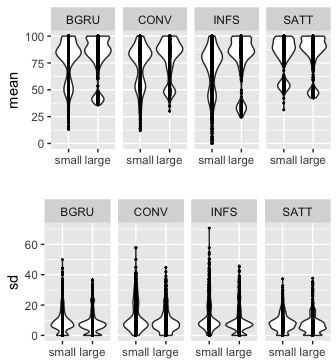}
	\caption{Performance on the known word perturbation probe, small and large training conditions (see \textsection\ref{sect:training-setup}).}
\label{fig:transformations}
\end{figure}

Recall from \textsection\ref{sec:known_word_probe} that the known word
perturbation probe involves insertion, deletion, or substitution of a
trained closed-class word in a correctly classified jabberwocky-type
sentence pair. Figure~\ref{fig:transformations} plots the results of
this probe. Each point represents a \fm{perturbation type}---a group
of perturbed test items that share their before/after target perturbed
closed-class words and before/after relation pairs.  The upper plot
displays the mean accuracy of all perturbations, averaged
across blocks, and the lower plot displays the standard deviations
across blocks.

All models perform substantially worse than the holdout-evaluation on
at least some of the perturbations. In addition, the standard
deviation of accuracy between blocks is higher than the holdout
tests. As discussed in \textsection\ref{sec:known_word_probe}, low
accuracy on this probe indicates that closed-class words do not
maintain a consistent interpretation when paired with different
open-class words. Variance across blocks shows that under all models
the behavior of closed-class words is highly sensitive to the novel
words they appear with.

Performance is also susceptible to interference from sentence-level
features. For example, consider the perturbation which deletes a
post-modifier from a sentence pair in negation, yielding a pair in
cover relation. The self-attentive encoder performs perfectly when
this perturbation is applied to a premise ($100\% \pm 0.00\%$), but
not when applied to a hypothesis ($86.60\% \pm 18.08\%$). Similarly,
deleting the adjective \textit{red} from the hypothesis of a
forward-entailing pair results in an unrelated sentence pair
($84.79\% \pm 7.50\%$) or another forward-entailing pair
($92.32\%, \pm 3.60\%$) or an equality pair ($100\% \pm 0.00\%$). All
the possible perturbations we studied exhibit similarly inconsistent
performance.

\subsection{Analysis IV: Identical Open-Class Words  Probe}
\label{sec:lexidentity_analysis}

Recall that the identical open-class words probe consist of sentence
pairs where all open-class lexical items were identical.
Table~\ref{tbl:freebies-small} shows the accuracies for these probes,
trained on the small language.  Average accuracies across jabberwocky
blocks are reported together with standard deviations.

\begin{table}[ht]
\centering
\resizebox{\linewidth}{!}{
\begin{tabular}{lllll}
\toprule
Relation & BGRU & CONV & SATT & INFS\\\
 & {\small mean (sd)} & {\small mean (sd)} & {\small mean (sd)} &{\small mean (sd)} \\
\midrule
\rowcolor{gray!6}  $\#$ & 100 $\pm{0}$ & 100 $\pm{0}$ & 99.94 $\pm{0.26}$ & 99.67 $\pm{0.98}$\\
$^\wedge$ & 55.68 $\pm{20.29}$ & 73.29 $\pm{10.8}$ & 23.71 $\pm{11.45}$ & 90.67 $\pm{10.98}$\\
\rowcolor{gray!6}  $\sqsubset$ & 90.78 $\pm{4.99}$ & 82.84 $\pm{6.51}$ & 75.22 $\pm{5.98}$ & 95.53 $\pm{2.64}$\\
$\equiv$ & 90.43 $\pm{17.1}$ & 38.12 $\pm{15.56}$ & 71.94 $\pm{24.1}$ & 95.93 $\pm{6.5}$\\
\rowcolor{gray!6}  $\sqsupset$ & 90.34 $\pm{4.18}$ & 77.11 $\pm{5.9}$ & 81.4 $\pm{6.67}$ & 93.81 $\pm{2.96}$\\
\addlinespace
$\mid$ & 93.08 $\pm{3.58}$ & 85.34 $\pm{5.47}$ & 74.05 $\pm{8.03}$ & 92.23 $\pm{4.6}$\\
\rowcolor{gray!6}  $\smile$ & 88.01 $\pm{3.55}$ & 71.5 $\pm{7.32}$ & 78.4 $\pm{7.91}$ & 95.22 $\pm{3.58}$\\
\bottomrule
\end{tabular}}
\caption{Identical open-class words probe performance, trained on the small language condition (trained on $51,743$ sentence pairs, see \textsection\ref{sect:training-setup})}\label{tbl:freebies-small}
\end{table}

Accuracy on the probe pairs fails to reach the holdout test levels for
most models and most relations besides $\#$, and variance between
blocks is much higher than in the holdout evaluation. Of special
interest is negation ($^\wedge$), for which accuracy is dramatically
lower and variance dramatically higher than the holdout evaluation.

The results are similar for the large language condition, shown in
Table~\ref{tbl:freebies-large}. Although model accuracies improve
somewhat, variance remains higher than the heldout level and accuracy
lower. Recall that these probe-items can be classified while ignoring
the specific identity of their open-class words. Thus, the models
inability to leverage this fact, and high variance across different
sets novel open-class words, illustrates their sensitivity to context.

\begin{table}[ht]
\centering
\resizebox{\linewidth}{!}{
\begin{tabular}{lllll}
\toprule
Relation & BGRU & CONV & SATT & INFS\\\
 & {\small mean (sd)} & {\small mean (sd)} & {\small mean (sd)} &{\small mean (sd)} \\
\midrule
\rowcolor{gray!6}  $\#$ & 99.82 $\pm{0.45}$ & 99.57 $\pm{0.73}$ & 98.67 $\pm{1.81}$ & 100 $\pm{0}$\\
$^\wedge$ & 84.18 $\pm{12.29}$ & 73.73 $\pm{18.31}$ & 79.97 $\pm{16.58}$ & 85.54 $\pm{14.11}$\\
\rowcolor{gray!6}  $\sqsubset$ & 96.13 $\pm{2.59}$ & 93.88 $\pm{2.67}$ & 97.3 $\pm{2.36}$ & 97.02 $\pm{2.39}$\\
$\equiv$ & 89.33 $\pm{12.5}$ & 77.84 $\pm{12.08}$ & 94.44 $\pm{11.23}$ & 94.59 $\pm{7.02}$\\
\rowcolor{gray!6}  $\sqsupset$ & 95.4 $\pm{2.48}$ & 94.55 $\pm{2.04}$ & 98.05 $\pm{1.51}$ & 97.6 $\pm{2.08}$\\
\addlinespace
$\mid$ & 89.97 $\pm{6.73}$ & 92.36 $\pm{6.29}$ & 84.52 $\pm{7.07}$ & 98.72 $\pm{2.08}$\\
\rowcolor{gray!6}  $\smile$ & 90.78 $\pm{6.33}$ & 93.18 $\pm{2.95}$ & 87.85 $\pm{6.46}$ & 97.48 $\pm{2.56}$\\
\bottomrule
\end{tabular}}
\caption{Identical open-class words probe performance when trained on the large language training condition (trained on $478,649$ sentence pairs, see \textsection\ref{sect:training-setup})}\label{tbl:freebies-large}
\end{table}

\subsection{Analysis V: Consistency Probe}
\label{sec:inconsistencies}

The consistency probe tests abstract knowledge of relationships
between logical relations, such as the fact that two sentences that
stand in a contradiction still stand in a contradiction after
reversing their order. Results of this probe in the small-language
condition are in Table~\ref{tbl:flipflop-small}: For each type of
relation, we show the average percentage of correctly-labeled sentence
pairs that, when presented in reverse order, were also correctly
labeled.

The best-performing model on negation reversal is SATT, which
correctly labeled reversed items $66.92\%$ of the time.  Although
performance on negation is notably more difficult than the other
relations, every model, on every relation, exhibited inter-block
variance higher than that of the hold-out evaluations.

\begin{table}[ht]
\centering
\resizebox{\linewidth}{!}{
\begin{tabular}{lllll}
\toprule
Relation & BGRU & CONV & SATT & INFS \\\
 & {\small mean (sd)} & {\small mean (sd)} & {\small mean (sd)} &{\small mean (sd)} \\
\midrule
\rowcolor{gray!6}  $\#$ & 97.4 $\pm{0.86}$ & 97.8 $\pm{0.93}$ & 98.58 $\pm{0.74}$ & 97.03 $\pm{0.87}$\\
$^\wedge$ & 63.03 $\pm{36.19}$ & 63.42 $\pm{35.91}$ & 66.92 $\pm{31.45}$ & 57.16 $\pm{38.24}$\\
\rowcolor{gray!6}  $\sqsubset$  & 92.45 $\pm{6.26}$ & 88.1 $\pm{8.16}$ & 93.16 $\pm{5.42}$ & 90.64 $\pm{6.76}$\\
$\equiv$ & 100 $\pm{0}$ & 100 $\pm{0}$ & 100 $\pm{0}$ & 100 $\pm{0}$\\
\rowcolor{gray!6}  $\sqsupset$ & 91.37 $\pm{6.23}$ & 94.73 $\pm{6.51}$ & 96.42 $\pm{3.22}$ & 87.02 $\pm{9.61}$\\
\addlinespace
$\mid$ & 96.02 $\pm{2.6}$ & 96.29 $\pm{2.51}$ & 96.95 $\pm{2.14}$ & 94.2 $\pm{3.48}$\\
\rowcolor{gray!6}  $\smile$ & 93.57 $\pm{3.56}$ & 95 $\pm{2.97}$ & 96.4 $\pm{2.83}$ & 93.1 $\pm{3.77}$\\
\bottomrule
\end{tabular}}
\caption{Consistency probe performance, trained on the small language condition ($51,743$ sentence pairs, see \textsection\ref{sect:training-setup}).}\label{tbl:flipflop-small}
\end{table}

Furthermore, as can be seen in Table~\ref{tbl:flipflop-large}, the
large language condition yields little improvement. Negation pairs are
still well below the hold-out test threshold, still with a high degree
of variation. Variation remains high for many relations, which is
surprising because the means report accuracy on test items that were
chosen specifically because the same item, in a reverse order, was
already correctly labeled. Reversing the order of sentences causes the
model to misclassify the resulting pair, more often for some blocks
than others.

\begin{table}[ht]
\centering
\resizebox{\linewidth}{!}{
\begin{tabular}{lllll}
\toprule
Relation & BGRU & CONV & SATT & INFS \\\
 & {\small mean (sd)} & {\small mean (sd)} & {\small mean (sd)} &{\small mean (sd)} \\
\midrule
\rowcolor{gray!6}  $\#$ & 98.45 $\pm{0.65}$ & 98.69 $\pm{0.54}$ & 98.83 $\pm{0.6}$ & 98.38 $\pm{0.74}$\\
$^\wedge$ & 70.46 $\pm{33.72}$ & 77.82 $\pm{26}$ & 84.27 $\pm{23.89}$ & 65.64 $\pm{35.13}$\\
\rowcolor{gray!6}  $\sqsubset$ & 96.02 $\pm{2.96}$ & 96.6 $\pm{3.26}$ & 96.78 $\pm{4.23}$ & 95.01 $\pm{5.38}$\\
= & 100 $\pm{0}$ & 100 $\pm{0}$ & 100 $\pm{0}$ & 100 $\pm{0}$\\
\rowcolor{gray!6}  $\sqsupset$ & 93.5 $\pm{4.51}$ & 95.76 $\pm{4.23}$ & 94.23 $\pm{5.86}$ & 90.11 $\pm{8.5}$\\
\addlinespace
$\mid$ & 96.31 $\pm{2.73}$ & 97.25 $\pm{2.05}$ & 97.17 $\pm{2.23}$ & 94.46 $\pm{4.24}$\\
\rowcolor{gray!6}  $\smile$ & 96.25 $\pm{2.49}$ & 96.98 $\pm{2.66}$ & 97.18 $\pm{2.17}$ & 93.88 $\pm{4.78}$\\
\bottomrule
\end{tabular}}
\caption{Consistency probe performance, trained on the large langauge condition ($478,649$ sentence pairs).}\label{tbl:flipflop-large}
\end{table}


\section{Discussion and Conclusion}

Systematicity refers to the property of natural language
representations whereby words (and other units or grammatical
operations) have consistent meanings across different contexts. Our
probes test whether deep learning systems learn to represent
linguistic units systematically in the natural language inference
task.  Our results indicate that despite their high overall
performance, these models tend to generalize in ways that allow the
meanings of individual words to vary in different contexts, even in an
artificial language where a totally systematic solution is available.
This suggests the networks lack a sufficient inductive bias to learn
systematic representations of words like quantifiers, which even in
natural language exhibit very little meaning variation.

Our analyses contain two ideas that may be useful for future studies
of systematicity.  First, two of our probes (known word perturbation
and consistency) are based on the idea of starting from a test item
that is classified correctly, and applying a transformation that should
result in a classifiable item (for a model that represents word meaning
systematically). Second, our analyses made critical use of differential
sensitivity (i.e., variance) of the models across test blocks with
different novel words but otherwise identical information content.  We
believe these are a novel ideas that can be employed in future studies.

\section*{Acknowledgements}
We thank Brendan Lake, Marco Baroni, Adina Williams,
Dima Bahdanau, Sam Gershman, Ishita Dasgupta, Alessandro Sordoni, Will Hamilton, Leon Bergen, the Montreal
Computational and Quantitative Linguistics, and Reasoning and
Learning Labs at McGill University for feedback on the manuscript. We
are grateful to Facebook AI Research for providing extensive
compute and other support. We also
gratefully acknowledge the support of the Natural Sciences and
Engineering Research Council of Canada, the Fonds de Recherche
du Qu\'{e}bec, Soci\'{e}t\'{e} et Culture, and the Canada
CIFAR AI Chairs Program.

\fi

\bibliography{acl2020}
\bibliographystyle{acl_natbib}

\end{document}